\def\year{2022}\relax

\documentclass[letterpaper]{article} 
\usepackage{aaai22}  
\usepackage{times}  
\usepackage{helvet}  
\usepackage{courier}  
\usepackage[hyphens]{url}  
\usepackage{graphicx} 
\usepackage{natbib}  
\usepackage{caption} 
\DeclareCaptionStyle{ruled}{labelfont=normalfont,labelsep=colon,strut=off} 
\graphicspath{{./figures/}}
\usepackage{amsmath}
\usepackage{amsfonts}
\usepackage{amssymb}
\usepackage{booktabs} 
\usepackage{subcaption} 
\usepackage{todonotes} 

\newcommand{\A}{\mathcal{A}}		

\DeclareMathOperator*{\argmax}{\textrm{argmax}}

\newcommand{\EE}[2][]{\mathbb{E}_{#1}\left[#2\right]}

\newcommand{\eg}{\textit{e.g.},~}
\newcommand{\Eg}{\textit{E.g.},~}
\newcommand{\ie}{\textit{i.e.},~}

\newcommand{\thetitle}{Global and Local Analysis of Interestingness for Competency-Aware Deep Reinforcement Learning}

\newcommand{\theauthors}{Pedro Sequeira, Jesse Hostetler, Melinda Gervasio}

\newcommand{\theabstract}{
In recent years, advances in deep learning have resulted in a plethora of successes in the use of reinforcement learning (RL) to solve complex sequential decision tasks with high-dimensional inputs. However, existing systems lack the necessary mechanisms to provide humans with a holistic view of their competence, presenting an impediment to their adoption, particularly in critical applications where the decisions an agent makes can have significant consequences. Yet, existing RL-based systems are essentially competency-unaware in that they lack the necessary interpretation mechanisms to allow human operators to have an insightful, holistic view of their competency. In this paper, we extend a recently-proposed framework for explainable RL that is based on analyses of ``interestingness.'' Our new framework provides various measures of RL agent competence stemming from interestingness analysis and is applicable to a wide range of RL algorithms. We also propose novel mechanisms for assessing RL agents' competencies that: 1) identify agent behavior patterns and competency-controlling conditions by clustering agent behavior traces solely using interestingness data; and 2) identify the task elements mostly responsible for an agent's behavior, as measured through interestingness, by performing global and local analyses using SHAP values. Overall, our tools provide insights about RL agent competence, both their capabilities and limitations, enabling users to make more informed decisions about interventions, additional training, and other interactions in collaborative human-machine settings.
}

\title{\thetitle}
\author {\theauthors}
\affiliations {
    SRI International\\
    333 Ravenswood Ave., Menlo Park, 94025 CA\\
    pedro.sequeira@sri.com, jesse.hostetler@sri.com, melinda.gervasio@sri.com
}

\begin{document}

\maketitle

\begin{abstract}
\theabstract
\end{abstract}

\section{Introduction}%
\label{Sec:Intro}



\begin{figure*}[!ht]
    \centering
    \includegraphics[width=\textwidth]{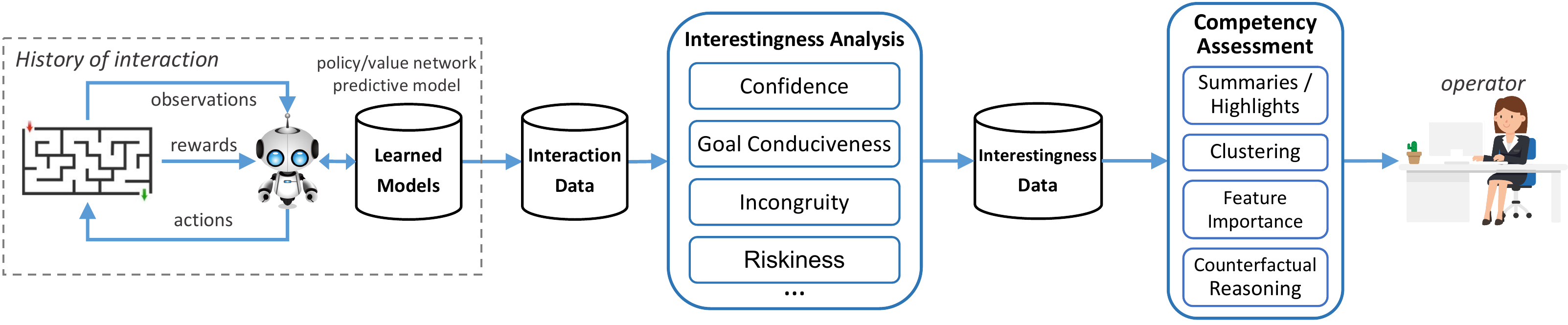}
    \caption{Our framework for analyzing the competence of deep RL agents through interestingness analysis.}%
	\label{Fig:Framework}
\end{figure*}

Reinforcement Learning (RL) is a machine learning technique for training autonomous agents to perform complex tasks through trial and error interactions with dynamic and uncertain environments. Recently, deep RL has achieved phenomenal successes, allowing agents to achieve---and even surpass---the performance level of human experts on various tasks, \cite[\eg][]{silver2018rl,vinyals2019alphastar,berner2019dota2}. In addition to solving complex tasks in simulated environments, RL has also been applied in real-life, industrial settings \cite[see][for a recent survey]{naeem2020rlapps}. However, an impediment to the wider adoption of RL techniques for autonomous control, especially in critical settings, is that deep learning-based models are essentially black boxes, making it hard to assess their competency in a task, and identify and understand the conditions that affect their behavior.

In deciding whether to delegate a task to an autonomous agent, a human needs to know that the agent is capable of making the right decisions under the various conditions to adequately accomplish the task. The challenge with an RL agent is that after being trained and deployed, it will always select one action at each step, as informed by its policy---but the why (and why not) behind decision making cannot be retrieved from the agent's model.
And, while we can test RL agents prior to deployment and gather statistics about their performance in the task, or identify the actions they will select under certain circumstances, a more complete understanding of agents' competence---both in terms of its capabilities and limitations---will be essential to their acceptance by human collaborators.

Instead of capturing agent competence in terms of the specific decisions it makes under certain circumstances or its performance along particular external metrics, we seek to characterize competence through self-assessment (introspection) over its history of interaction with the environment to capture distinct aspects that help explain its behavior. 
To do that, we analyze trained RL policies along various \emph{interestingness} dimensions \cite{sequeira2019interestingness}, \ie information that has the potential to be ``interesting'' in helping humans understand the competence of an RL agent.
The interestingness analyses capture aspects of agent competence such as whether an agent is confident in selecting actions, or whether it recognizes risky or unfamiliar situations, among other things. In general, the goal is to direct end-users of the RL system towards appropriate intervention by providing deeper insight into the agent's competence, \eg identifying sub-task competency (the situations in which the agent is more/less competent), or highlighting the situations requiring more training or direct guidance.

In this paper, we investigate the varied forms by which we can use interestingness analyses to better understand the competency of deep RL agents. The focus is on the quantitative and qualitative analyses over interestingness data, and on the insights about an agent's behavior that our framework can provide to potential end-users of our system. 
Fig.~\ref{Fig:Framework} depicts our framework for competency-aware deep RL through interestingness. The input to our system is a trained deep RL policy that provides a set of \emph{learned models} that depends on the underlying RL algorithm, but typically corresponds to policy and value deep neural networks. 
Then, we test the agent by deploying it in the environment a number of times under different initial conditions, resulting in a set of traces. As the agent interacts with the environment, we probe the learned models and collect various information about the agent's behavior and internal state, \eg the value attributed to a state, or the probability each action has of being selected---we refer to all this information as the \emph{interaction data}.
Then, we perform interestingness analysis through introspection along several dimensions, resulting in a scalar value for each timestep of each trace---the \emph{interestingness data}. 
Finally, we perform competency assessment based on interestingness using diverse techniques. The use of interestingness for RL behavior summarization and counterfactual reasoning are covered respectively in \cite{sequeira2020ixrl} and \cite{yeh2022counterfactuals}; here we focus on the remaining two mechanisms.
%
Our contributions are as follows:
\begin{itemize}
    \item A new set of interestingness dimensions designed to cover various families of deep RL algorithms.%
    \footnote{Our previous system was compatible only with tabular RL.}
    Our implementation of the interestingness framework is compatible with popular RL toolkits.%
    \footnote{The code with our implementation will be posted online.}
    \item A method to analyze an RL agent's behavior in a task by clustering traces based solely on interestingness, that allows isolating distinct competency-inducing conditions resulting in different behavior patterns.
    \item A method to discover which task elements impact agent’s behavior the most and under which circumstances. We conduct feature importance analysis via SHapley Additive exPlanations (SHAP) \cite{lundberg2017shap} to perform global and local interpretation for competency assessment.
\end{itemize}
%
We present the results of a computational study where we trained an RL policy in a combat task running on the StarCraft II (SC2) platform and then utilized our interestingness framework to analyze the resulting agent's behavior and competence in the task. 
We show that by going beyond trying to capture what an agent will do when and how well it will do it, our interestingness dimensions can give human operators and partners a more complete picture of an RL agent's competence.
Furthermore, the trace clusters (obtained based on interestingness data) expose disparate challenges resulting in distinct agent behaviors, allowing the identification of different sub-tasks within the general SC2 task and distinct competency-controlling conditions. In addition, our feature importance analysis via SHAP provides insights about the ``sources'' of interestingness in the environment that most impact the agent’s competence, while also helping explain the contributions of each task element in particular situations having ``abnormal'' values of interestingness. Altogether, the higher-level competency modeling enabled by interestingness provides human users with a more complete understanding of an agent's competencies, allowing them to make better decisions regarding the agent's use, providing insights on the agent’s limitations in the task, and suggesting directions for improving agent behavior.

\section{Related Work}%
\label{Sec:RelatedWork}

In recent years, many approaches to eXplainable RL (XRL) have been proposed for explaining various aspects of learned policies \cite{heuillet2021xrl, puiutta2020xrl}. These include identifying the regions of the input that most affect an agent's decisions \cite{zahavy2016understanding,greydanus2018visualizing}, providing example trajectories \cite{huang2019enabling}, highlighting key decision moments to summarize an agent's behavior \cite{amir2019,huang2018establishing,lage2019summary}, extracting high-level descriptions of an agent's policy \cite{dereszynski2011learning,hostetler2012inferring,sequeira2022camly2,koul2019fsm}, and generating counterfactual explanations to help in understanding an agent's behavior \cite{madumal2020causal,yeh2022counterfactuals,olson2021counterfactual}. Here we focus on the approaches that attempt to provide a more holistic view of RL agent competence in a form understandable to humans. 

One body of approaches focuses on explaining behavior in the form of queryable models that let users analyze agent behavior under distinct conditions. For example, \citet{hayes2017policyexplanation} find correlations between conditions and actions and let users query the model through templated questions around identifying conditions for actions, predicting what an agent will do, and explaining expectation violations. Meanwhile, \citet{vanderwaa2018contrastive} use simulation to predict sequences of future actions to answer user questions about the consequences of actions/policy. 

Another body of work attempts to characterize agent behavior by extracting diverse types of structural models to represent agent strategies. For example, \cite{dereszynski2011learning,hostetler2012inferring} use probabilistic graphical modeling to learn finite state models of strategy for the StarCraft domain, while \citet{sequeira2022camly2} infer strategies in the form of logical task specifications from agent traces using information-theoretic techniques to capture the conditions under which distinct behaviors occur. These approaches are focused on characterizing agent behavior and do not attempt to capture other aspects of the agent's decision-making. 

In earlier work on interestingness analysis \cite{sequeira2020ixrl}, we conducted a user study that involved showing users video clips, highlighting different interestingness moments of agent interactions with the environment. The results revealed that some dimensions are better than others at conveying the agent's capabilities and limitations but that the diversity of aspects captured by the distinct interestingness dimensions helped users better understand an agent's task competencies.

\section{Interestingness Analyses}%
\label{Sec:Interestingness}

In this section we detail our framework for analyzing RL agent competence through interestingness, starting by providing the necessary background on RL.

\subsection{Reinforcement Learning}%
\label{Subsec:RL}

We are interested in characterizing the competence of RL agents \cite[we refer to][for an introduction on the problem and main approaches]{sutton2018rl}. 
%
The goal of an RL algorithm is to learn a policy, denoted by $\pi(a|s)$, mapping from states to actions, 
that maximizes the expected return for the agent, \ie the discounted sum of rewards it receives during its lifespan. The optimization problem can be formulated by $\pi^*=\argmax_\pi{\EE{\sum_t{\gamma^t R_t}}}$, where $R_t$ is the reward received by the agent at discrete timestep $t$, and $\pi^*$ is termed the optimal policy. Often, RL algorithms use an auxiliary structure while learning a policy called the value function, corresponding to $V^\pi(s)=\EE{\sum_t{\gamma^t R_t})|S_0=s}$, that provides an estimate of the return the agent will receive by being in state $s$ and following policy $\pi$ thereafter, where $\gamma\in[0,1]$ is a discount factor denoting the importance of future rewards. In deep RL, policies and other auxiliary structures are represented by neural networks whose parameters can be adjusted during training to change the agent's behavior via some optimization algorithm. As indicated in Fig.~\ref{Fig:Framework}, we refer to all such networks optimized via RL as the \emph{learned models}.

\subsection{Interaction Data}%
\label{Subsec:InteractionData}

The next step in our framework is the extraction of \emph{interaction data} given a trained RL agent. We collect these data by ``running'' the agent in the environment for a number of times%
\footnote{Without loss of generality, here we deal with episodic tasks.}
and collecting samples from the environment and by probing the learned models at each step. The result is a set of \emph{traces} comprising the agent's history of interaction with the environment from which the agent's competence is going to be analyzed.

Distinct RL algorithms make use of different models to optimize the agent's policy during training. Our framework extracts data mainly from two separate families of RL algorithms, namely:%
\footnote{Due to space restrictions, the interaction data and interestingness dimensions reported here pertain only to the type of RL algorithm used in our SC2 experiments. Our full framework makes use of data provided by RL algorithms from other families, including model-based and distributional RL approaches, allowing us to compute interestingness along additional dimensions of analysis.}
\begin{description}
    \item [Policy gradient] approaches \cite[\eg][]{sutton2018rl,schulman2017ppo,haarnoja2018sac} that provide a stochastic policy, $\pi(a|s)$, \ie a function mapping from observations to distributions over the agent's actions; 
    \item[Value-based] approaches \cite[\eg][]{mnih2015dqn,schaul2015replay} that compute a (state) value function, $V(s)$, that indicates how good it is for the agent to be in a situation, or an action-value function, $Q(s,a)$, asserting the value of executing some action given a state;
    %
    %
    %
\end{description}

Interaction data then comprises everything we can extract given the learned models provided by the RL agent at each timestep. In addition to this internal agent data, we also collect external (observable) data, \ie the reward received by the agent, the selected action and the agent's observation.

\subsection{Interestingness Dimensions}%
\label{Subsec:Dimensions}

As mentioned earlier, the goal of interestingness analysis is to characterize RL agents' competence along various dimensions, each capturing a distinct aspect of the agent's interaction with the environment. The dimensions of analyses are inspired by what humans---whether operators or teammates---might seek when trying to understand an agent’s competence in a task. Each dimension provides distinct \emph{targets of curiosity} whose values might trigger a human to investigate the agent's learned policy further \cite{hoffman2018xai}. As such, our system provides a means for the agent to perform competency self-assessment, where we use the data resulting from the interestingness analyses to identify cases that a human should be made aware of, and where user input might be needed.

Computationally, an analysis is given interaction data for each trace as input (see Fig.~\ref{Fig:Framework}), and, for each timestep, produces a scalar value in the $[-1,1]$ interval denoting the ``strength'' or ``amount'' of competence as measured by that interestingness dimension. Further, with the goal of allowing the analyses to be computed online, \ie \emph{while} the agent is performing the task, we restrict analyses to have access only to the data provided up to a given timestep. 
%
In this work, we designed and implemented a novel set of interestingness analyses that are applicable to a wide range of tasks and cover most of existing deep RL algorithms.%
\footnote{Notwithstanding, different RL algorithms allow the collection of only a subset of the interaction data mentioned in Sec.~\ref{Subsec:InteractionData}, which will result in a subset of the interestingness analyses being performed given a particular RL agent.}
Moreover, our implementation is compatible with popular RL toolkits, including RLLib \cite{liang2018rllib}, an open-source library offering support for production-level, highly distributed RL workloads (20+ RL algorithms), which can foster wider adoption of our tool.

We now describe the goal behind each interestingness analysis. We detail only the analyses and their realization that are compatible with the type of state and action space and RL algorithm used in our SC2 experiments. 
\begin{description}
    \item[Value:] characterizes the long-term importance of a situation as ascribed by the agent’s value function. It can be used to identify situations where the agent nears its goals (maximal value) or in which it is far from them (low value). Given the agent’s value function associated with policy $\pi$, denoted by $V^{\pi}$, we compute \emph{Value} at discrete timestep $t$ using: $\mathcal{V}(t)=2\left(V_{[0,1]}^\pi(s_t)\right)-1$, where $V_{[0,1]}$ is the normalized value function obtained via min-max scaling across all timesteps of all traces.
    \item[Confidence:] measures how confident an agent is with regards to action-selection, helping identify good opportunities for requesting guidance from a human user or 
    sub-tasks in which the agent would require more training. In our experiments, we consider discrete action spaces, where a stochastic RL policy is applicable, corresponding to a discrete probability distribution over the possible actions. As such, we compute \emph{Confidence} at timestep $t$ with $\mathcal{C}(t)=1-2J\left(\pi(\cdot|s_t)\right)$, where $J(X)$ is Pielou’s evenness index \cite{pielou1966evenness}, corresponding to the normalized entropy of a discrete distribution $X$, which is given by $J(X)=-\frac{1}{\log{n}}\sum_i{P(x_i)\log{P(x_i)}}$.
	\item[Goal Conduciveness:] assesses how desirable for the agent a situation is given the context of the decision at that point, \ie the preceding timesteps leading up to the current state. Intuitively, it computes how ``fast'' the agent is moving towards or away from the goal. Decreasing values can be particularly interesting, but we can also capture large differences in values, potentially identifying external, unexpected events that would violate operator's expectations and in which further inspection would be required. We compute \emph{Goal Conduciveness} at timestep $t$ directly from the first derivative of the value function with respect to time at $t$, namely using: $\mathcal{G}(t)=\sin\left(\arctan\left(\rho \frac{d}{dt}V_{[0,1]}(s_t) \right)\right)$, where the sine of the angle generated by the slope (in an imaginary unit circle centered at $V_{[0,1]}(s_t)$) is used for normalization and $\rho$ is a scaling factor to make the slope more prominent. In our implementation, we use $\rho=100$, and resort to a finite difference numerical method to approximate $\frac{d}{dt}V(s_t)$ given the value function of the $3$ previous timesteps.%
	\footnote{This corresponds to using the \emph{backward} finite difference coefficient with accuracy $2$ \cite{fornberg1988findiff}. A higher-order accuracy could be used if we wish to capture how the value function is changing for the computation of Goal Conduciveness, by using information from timesteps further back in the trace.}
	\item[Incongruity:] captures internal inconsistencies with the expected value of a situation, which might be useful to indicate unexpected situations, \eg where reward is stochastic or very different than the one experienced during training. In turn, a prediction violation identified through incongruity can be used to alert a human operator about possible deviations from the expected course of action. Formally, we capture \emph{Incongruity} via the temporal difference (TD) error \cite{sutton2018rl}, \ie $\mathcal{I}(t)=r_t+\gamma V^\pi(s_t)-V^\pi(s_{t-1})$.%
	\footnote{This quantity is also known as the \emph{one-step TD} or \emph{TD(0) target}.}
	We then normalize $\mathcal{I}(t)$ by dividing it with the reward range observed from the task. 
	\item[Riskiness:] quantifies the impact of the ``worst case scenario'' at each step, highlighting situations where performing the ``right'' vs. the ``wrong'' action can dramatically impact the agent's long-term objectives. This dimension is best computed for value-based RL algorithms by taking the difference between the value of the best action, $\max_{a\in\A}{Q(a|s_t)}$ and the worst, $\min_{a\in\A}{Q(a|s_t)}$. However, here we use a policy-gradient algorithm that updates the policy directly. As such, we compute \emph{Riskiness} using $\mathcal{R}(t)=2\left(\max_{a_1\in\A}{\pi(a_1|s_t)} - \max_{a_2\in\A}{\pi(a_2|s_t)} \right)-1, a_1 \neq a_2$. This will usually result in a value similar to that of Confidence, but might help to identify situations where \emph{one} of the actions is particularly undesirable (low probability) compared to all the others, which can be used by an operator to further specify the conditions in which an action should never be executed.
\end{description}

	
\section{Experiments and Results}%
\label{Sec:Experiments}

To validate our interestingness framework and understand the insights it can provide to potential end-users of the system, we performed a computational study where we extracted interestingness data and applied distinct methods of interpreting the agent's competency in a combat scenario.%
\footnote{Although we do not perform a user study, our experiments are a necessary first step to identify the types of insights about RL agent competence that the interestingness analyses can provide.}

\subsection{StarCraft II Environment}%
\label{Subsec:Task}

\begin{figure}[!tb]
    \centering
    \includegraphics[width=0.9\columnwidth]{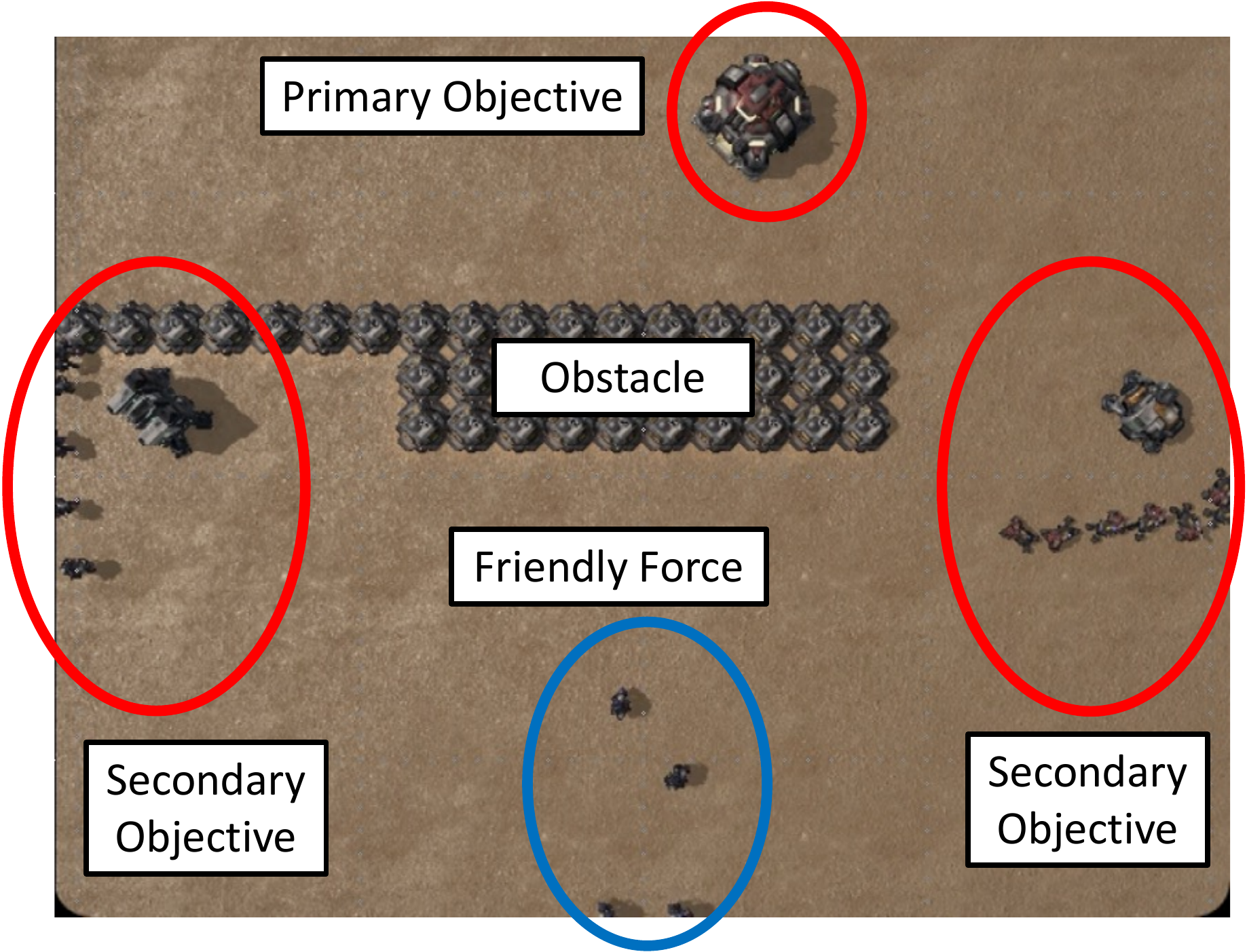}
    \caption{The SC2 combat task used in our experiments.}%
	\label{Fig:SC2Scenario}
\end{figure}

We implemented an RL task in the video game \textit{Starcraft~II} \citep[SC2,][]{blizzard2022starcraft} using the \texttt{pySC2} library \citep{vinyals2017starcraft} to interface with the game engine. In our SC2 task (an example of which is depicted in Fig.~\ref{Fig:SC2Scenario}), the agent controls the Blue force, which starts at the bottom of the map. The agent's goal is to destroy the Primary Objective, which is a CommandCenter (CC) building that appears at the top of the map. The map is divided into three vertical ``lanes,'' each of which may be blocked by obstacles. The two side lanes contain Secondary Objectives, which are buildings guarded by Red forces. Destroying the Red force defending one of the secondary objectives causes the building to be removed from the map and replaced with additional Blue units (reinforcements). The type of the reinforcements is determined by the type of the building. Different types of units have distinct capabilities. The starting Blue force consists of infantry (Marines or Marauders), but Blue can gain access to SiegeTanks (armored ground units) by capturing a Factory building, or to Banshees (ground attack aircraft) by capturing a Starport. The Banshees are especially important because they can circumvent obstacles and Red has no anti-air units defending the primary objective. 

The initial state, including the number and type of Blue starting units, locations of obstacles, number of enemies, etc. is randomized to create a family of \emph{scenarios}.%
\footnote{In addition, Red sometimes receives reinforcements at a random time step, and capturing a secondary objective sometimes does not grant Blue reinforcements.} 
The reward is a linear combination of five factors: a large reward for capturing the CC, rewards for gaining and losing Blue forces and for destroying Red forces in proportion to the resource cost of the units gained or lost, and a per-timestep cost.

The agent observes the world as a top-down grid view similar to a multi-channel image of size $192\times144$ pixels, each corresponding to a location on the map, and each channel encodes a distinct property of the object at that location. 
The action space is factored over the four unit types that Blue can have. A joint action assigns an order separately to each unit type, and all units of that type execute the same order. The available orders are to \emph{target} the nearest Red unit of a specified type, 
to \emph{move} to one of $9$ fixed locations while ignoring any enemies encountered, or to \emph{do nothing}. 

\subsection{RL Agent}%
\label{Subsec:Agent}

We trained a deep RL agent for our StarCraft task using a distributed implementation of the VTrace algorithm \citep{espeholt2018impala} based on the open-source SEED-RL codebase \citep{espeholt2020seedrl}. 
%
%
As VTrace is an actor-critic method, it uses both a policy network that outputs action distributions and a value network that estimates the state value. 
This allows us to extract interestingness using all five dimensions listed in Sec.~\ref{Subsec:Dimensions}. Furthermore, because in our task the agent controls $4$ different types of units and the action space is discrete, 
our RL policy outputs $4$ categorical distributions. 
As a result, Confidence and Riskiness can be computed for each unit type separately. 
We also compute the mean value across unit types, resulting in a total of $13$ interestingness variables that can be extracted from our RL agent.

\subsection{Agent Performance Results}%
\label{Subsec:PerformanceResults}

After training an RL agent in the SC2 task, we began analyzing the agent's competency using our interestingness framework. We started by sampling $1{,}000$ traces by running the trained policy on the SC2 task on scenarios sampled randomly. 
The mean trace length is $160.00\pm 81.36$ timesteps. Performance-wise, the agent achieved its primary objective (destroying the CC) $36\%$ of time, gets defeated (no Blue units left) in $25\%$ of traces while $39\%$ of traces timed out (the agent did not achieve the main goal but was also not defeated). The mean count of agent (Blue) units is $4.23\pm 3.47$ while the count for opponent (Red) units is $13.34\pm 5.59$. 

Overall, the performance of our RL agent is not optimal in that the agent did not destroy the CC in all traces, nor did it consistently destroy the Red units ($15\%$ of the number of initial units, on average). However, given the characteristics of our SC2 task, where achieving the main goal depends on the units the agent is given initially and the scenario configuration, \eg whether a Starport is present, the agent still achieves good performance. The results indicate that there appear to be task conditions for which the agent learned a strategy capable of achieving the task objectives, whereas other situations pose challenges that the agent cannot overcome, and in which further training or guidance might be needed. This makes the agent an ideal candidate for interestingness analysis of competence.

\subsection{Interaction Data and Interestingness Analyses}%
\label{Subsec:InterestingnessResults}


\begin{figure*}[!htb]
    \centering
    \begin{subfigure}[b]{0.33\textwidth}
        \includegraphics[width=\textwidth]{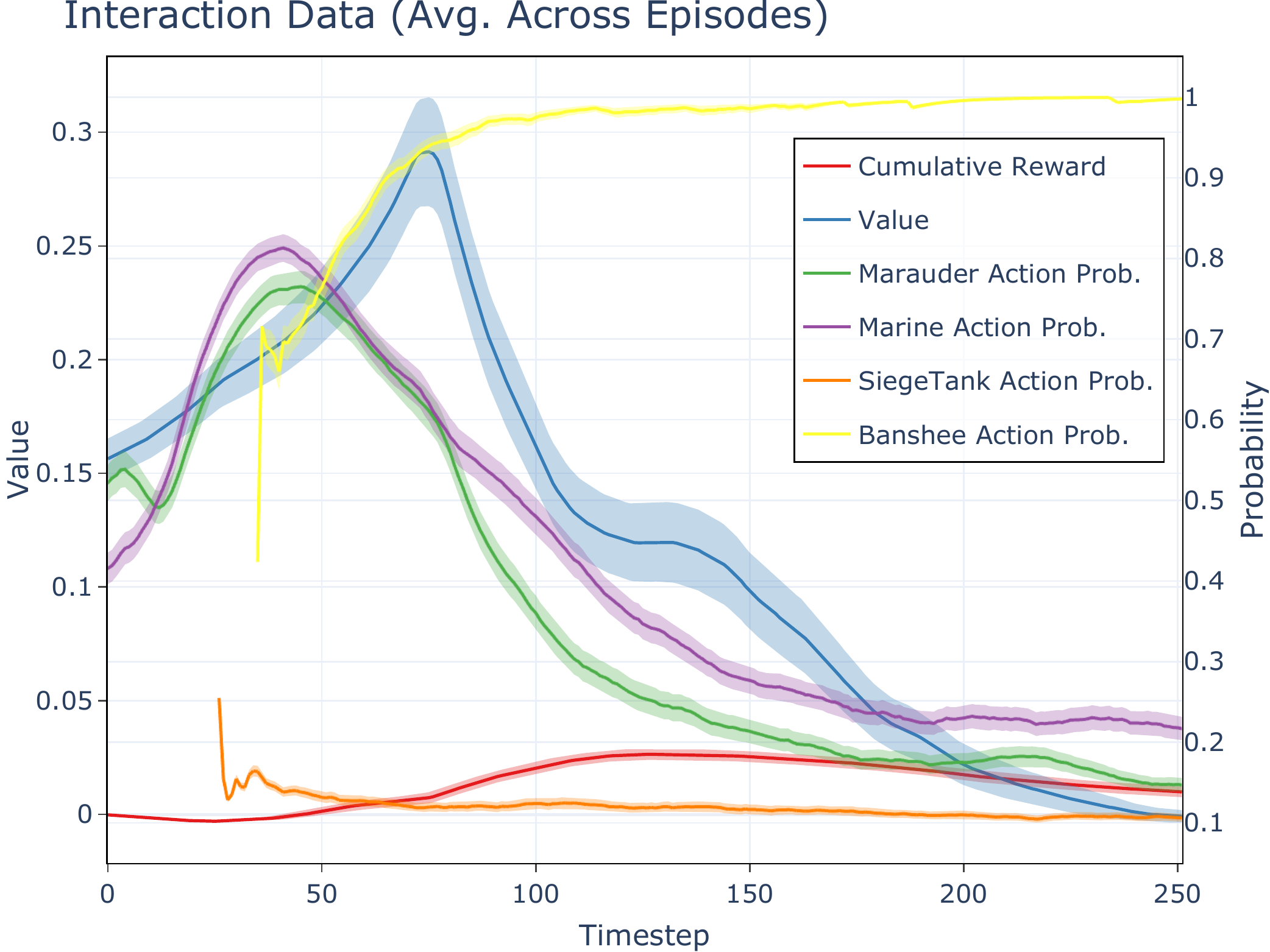}
        \caption{Mean interaction data.}%
	    \label{Fig:InteractionData}
    \end{subfigure}%
    \begin{subfigure}[b]{0.33\textwidth}
        \includegraphics[width=\textwidth]{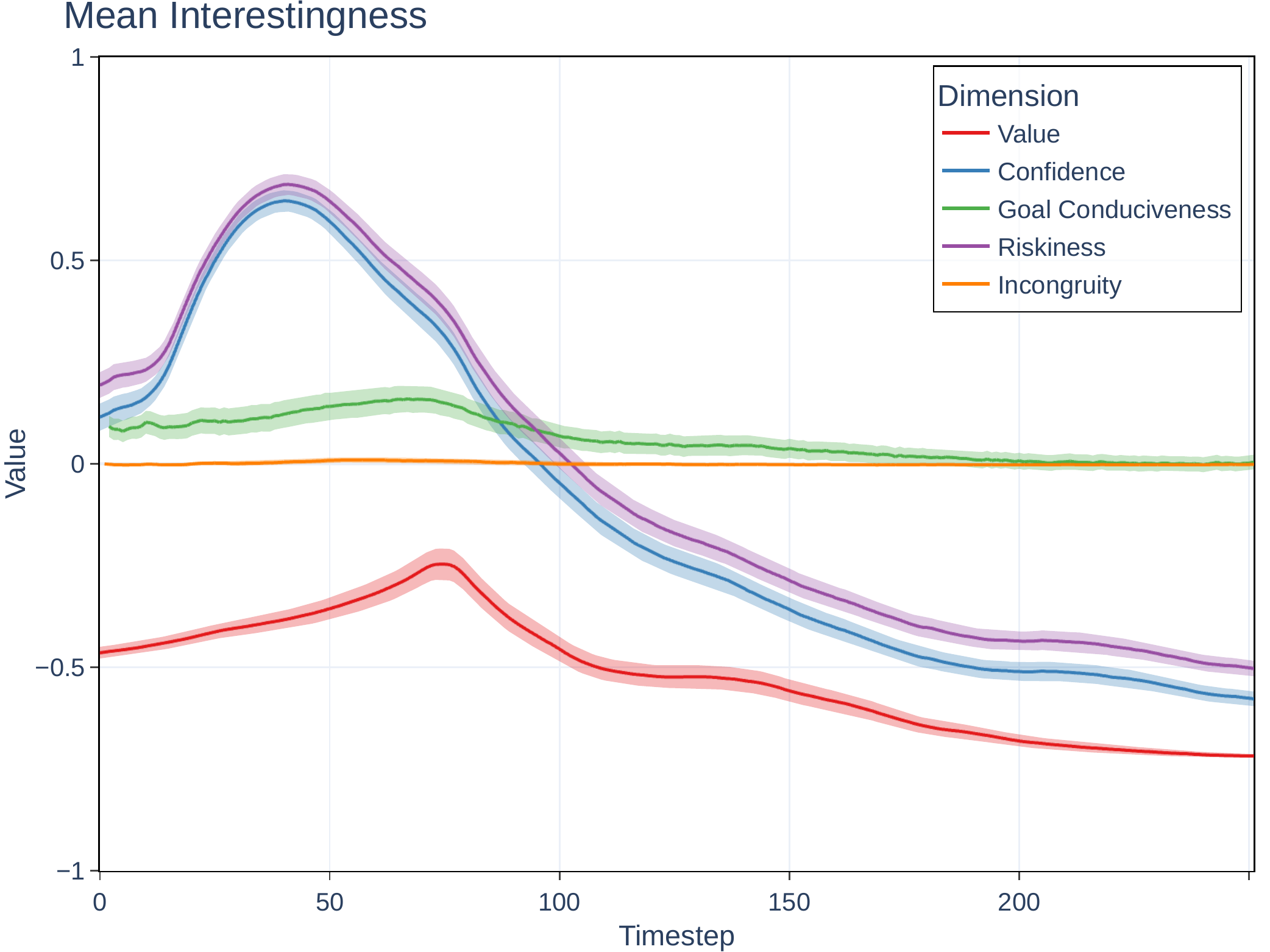}
        \caption{Mean interestingness over time.}%
        \label{Fig:MeanInt}
    \end{subfigure}%
    \begin{subfigure}[b]{0.33\textwidth}
        \includegraphics[width=\textwidth]{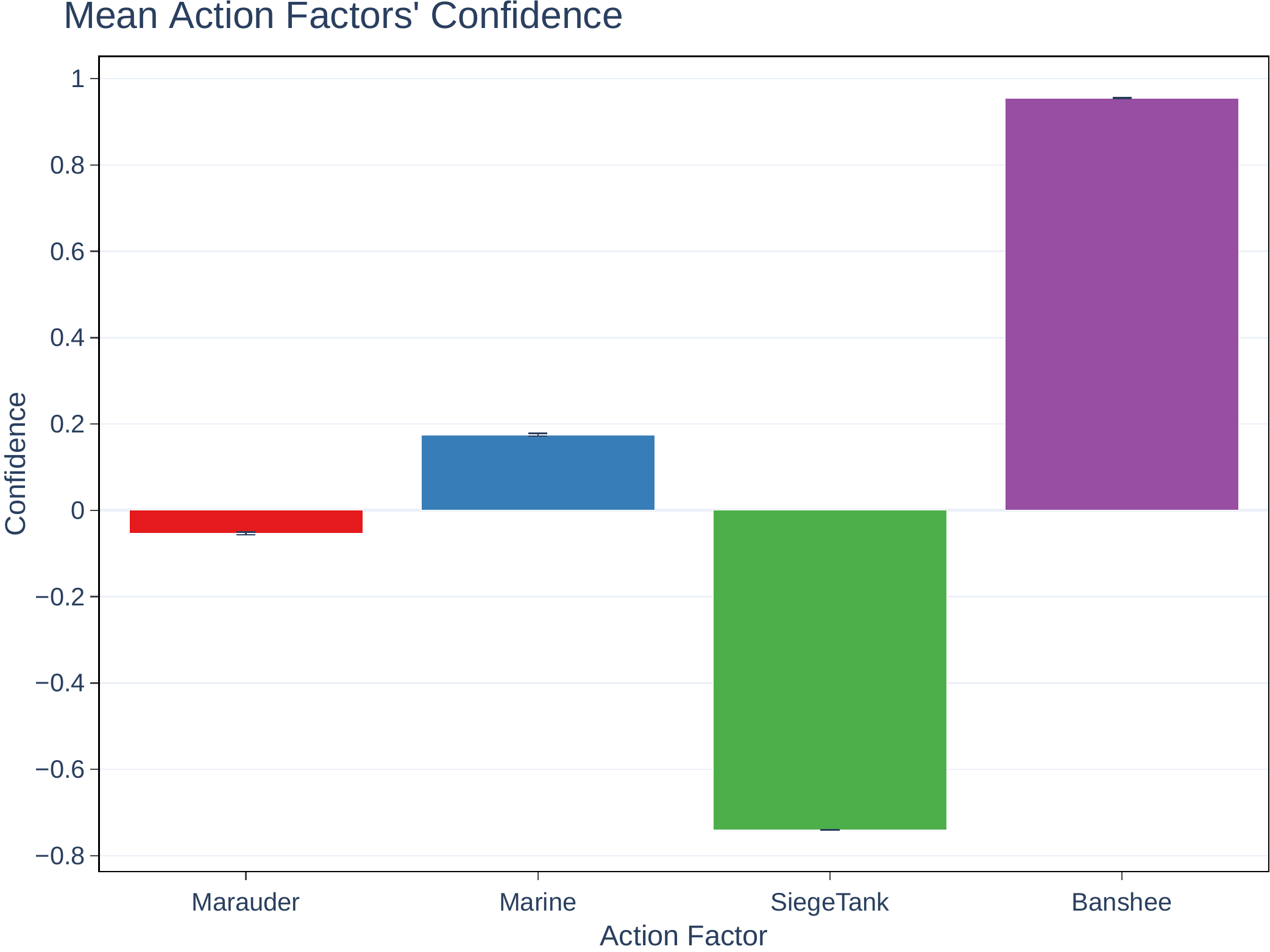}
        \caption{Mean confidence for unit types.}%
        \label{Fig:MeanConf}
    \end{subfigure}%
    \caption{Overview of interestingness analysis results over $1{,}000$ traces. Shaded regions and error bars represent the $95\%$ CI.}%
	\label{Fig:IntResults}
\end{figure*}

After sampling the behavior traces, we collected interaction data as explained in Sec.~\ref{Subsec:InteractionData}. Fig.~\ref{Fig:InteractionData} shows the mean interaction data over the course of a trace, averaged across all $1{,}000$ traces. As we can see, the agent seems to achieve its objectives during the first half of traces---this is visible in the decrease of the cumulative reward and value after the initial phase, meaning that after that, on average the agent just wanders in the environment accumulating the per-step negative reward. Moreover, looking at the action selection probability (note the secondary y-axis) for different unit types enables us to identify how the agent tends to use its forces---\eg the agent is more decisive when assigning orders to Banshees (higher probability) compared to other units; In contrast, the agent is less decisive when using SiegeTanks. This is a first indication that the agent's competence might not be uniform across all scenarios---otherwise, value and cumulative reward would always increase over time---and that using different unit types might lead to disparate degrees of competence---otherwise, the probability of selecting actions for all unit types would be consistently high. Human operators looking solely at aggregate \emph{performance} statistics would miss these nuances about the agent's competence and be less able to identify the competency-controlling conditions. 

Still, interaction data alone is not enough to isolate those aspects of competence we are interested in. As such, we performed interestingness analysis using the aforementioned $13$ dimensions over the $1{,}000$ collected traces. 
%
Fig.~\ref{Fig:MeanInt} shows the average evolution of interestingness over the course of traces (we show only results for the five main dimensions, excluding the per-unit dimensions). We observe that, on average across all timesteps of all traces, Value 
slightly increases over time (positive but low Goal Conduciveness), and that the agent has decreasing Confidence in selecting actions as time goes by. In addition, the great majority of situations experienced during testing seem to be highly predictable since Incongruity is near $0$ at all times. Overall, we can see that interestingness varies over time (x-axis) and across traces (shaded error regions), which indicates that different traces have associated disparate interestingness patterns that might denote distinct behaviors and conditions.
%
Interestingly, Fig.~\ref{Fig:MeanConf} shows that the agent is more confident in the use of some types of units compared to others. \Eg it is highly confident in the use of the Banshees but not the SiegeTanks. This confirms our initial assessment, but we need a more careful analysis of the agent's competence to understand what factors influence it the most.

\subsection{Trace Clustering based on Interestingness}%
\label{Subsec:Clustering}

\begin{table*}[!ht]
    \centering
    \footnotesize
    \begin{tabular}{l|l}
    \toprule
        \textbf{ID} & \textbf{Description} \\
    \midrule
        0 & Marauders attack CC directly or sec. obj. but fail / get killed. \\ 
        1 & Marauders attack sec. obj., get SiegeTanks, then wander around. \\
        2 & Marines attack CC when unobstructed and unprotected, or attack sec. obj. but get killed. \\
        3 & Marines attack sec. obj., gets SiegeTanks; use Marines to attack other sec. obj. but gets killed; SiegeTanks wander around. \\ 
        4 & Marines attack Starport, get Banshees, use them to destroy CC; Marines sometimes go to bottom, otherwise attack CC. \\ 
        5 & Variant of 4 (only 3 traces) where all Marines are killed (after getting the Banshees). \\ 
        6 & Marines destroy both sec. obj., get SiegeTanks; then wander around. \\
    \bottomrule
    \end{tabular}
    \caption{Description of main events and agent behavior in each cluster discovered through interestingness.}%
    \label{Table:ClustersDesc}
\end{table*}

With the purpose of using interestingness to recognize distinct, meaningful patterns---which would indicate that the agent has achieved some level of competence---we clustered the agent's traces using only the interestingness data. Our main hypothesis is that each cluster will highlight different behaviors and represent distinct aspects of competence as captured by interestingness. 
%
To compute distances between all pairs of traces, we used Dynamic Time Warping (DTW) \cite{salvador2007dtw}, which measures the similarity between two sequences by matching each index of one sequence into one or more indices from the other. Since we have data for multiple dimensions, we used DTW with a Euclidean metric to compute the alignment cost at each timestep. The resulting pairwise distances were then fed to a Hierarchical Agglomerative Clustering \citep{kaufman1990agglomerative} algorithm with a complete linkage criterion. To select the number of clusters, we compute the Silhouette coefficient \citep{rousseeuw1987silhouette}. 

The best partition resulted in $8$ clusters; however one cluster had a single trace so here we analyze the results using the second-best partition, with $7$ clusters. Two major clusters emerged that separated the data into traces where the agent starts 
either with Marines or Marauders; 
this indicates that the type of units the agent starts with is a competency-controlling condition, leading to distinct behavioral patterns as captured by interestingness.


We also analyzed the main task events and agent's behavior in each resulting cluster, which are described in Table~\ref{Table:ClustersDesc}.%
%
\footnote{Videos of sampled traces and spatio-temporal patterns visualizations \citep[described in][]{sequeira2022camly2} for each cluster can be found at: \url{https://github.com/SRI-AIC/22-llaama-ixdrl-data}.}
Overall, we observe more distinctive behavior patterns when using Marines vs. Marauders. Moreover, the clusters denote diverse challenges leading to distinct agent behaviors and to different outcomes. The scenarios in some clusters are indicative of situations where agent behavior can be improved, while others highlight situations where the agent behaves optimally (achieves its objectives), thereby denoting additional competency-controlling conditions.

\begin{table}[!t]
    \centering
    \footnotesize
    \begin{tabular}{l | r r r r r}
    \toprule
        \textbf{ID} & \textbf{Value} & \textbf{Conf.} & \textbf{Goal Cond.} & \textbf{Riskiness} & \textbf{Incong.} \\
    \midrule
        0 & $-0.48$ & $0.03$ & $0.07$ & $0.11$ & $0.00$ \\
        1 & $-0.71$ & $-0.37$ & $0.00$ & $-0.27$ & $0.00$ \\
        2 & $-0.39$ & $0.31$ & $0.15$ & $0.35$ & $0.00$ \\ 
        3 & $-0.69$ & $-0.26$ & $0.00$ & $-0.13$ & $-0.01$ \\
        4 & $0.04$ & $0.36$ & $0.19$ & $0.36$ & $0.01$ \\ 
        5 & $0.00$ & $0.72$ & $0.13$ & $0.71$ & $0.00$ \\ 
        6 & $-0.65$ & $-0.25$ & $0.02$ & $-0.17$ & $0.00$ \\ 
    \bottomrule
    \end{tabular}
    \caption{Mean values of interestingness for each cluster.}
    \label{Table:ClustersInt}
\end{table}

Additionally, as shown in Table~\ref{Table:ClustersInt}, the clusters resulted in characteristically different interestingness ``profiles,'' each denoting distinct behavior patterns. Looking at Value, we see a clear distinction between traces in which the agent gets the Banshees (positive Value) and all other clusters (negative Value), indicating that the presence of Banshees is a competency-controlling condition. For Confidence, we observe that clusters in which the agent receives SiegeTanks lead to low Confidence in the use of \emph{other} types of units, reinforcing the idea that the agent did not learn well what to do with these units. Moreover, Goal Conduciveness clearly separated traces based on whether the agent achieved the main or secondary objectives, thus showing the usefulness of this dimension for competency awareness. Finally, Incongruity shows that in some clusters, the agent clearly over/underestimated the value of some situations or the received reward. Overall, these results on using interestingness to cluster traces show that the several dimensions capture distinctive aspects of the agent's competencies, resulting in clusters where the agent tackles different challenges and exhibits distinct strategies. 

\subsection{Feature Importance Analysis}%
\label{Subsec:FeatureImportance}

Feature importance analysis allows us to gain deeper insight into which task elements the agent's competence is more affected by, how they affect the agent's behavior as measured by interestingness, and where (in which situations) this occurs. 
%
To do that, we need a set of high-level, interpretable features whose influence on each dimension of analysis we can predict and explain. Given that the \texttt{pySC2} interface provides semantic but rather low-level information, we utilized the high-level SC2 feature extractor described in \cite{sequeira2022camly2}, which provides numeric descriptions for the task elements (types and properties of units, amount and size of forces, etc.) and abstracts the agent's local behavior (movement of groups of units relative to the opponent, the orders assigned to each unit type, etc.).

To identify which features impact each interestingness dimension the most, we used SHapley Additive exPlanations (SHAP) \cite{lundberg2017shap}. SHAP values provide explanations by computing the impact of each feature's value on the prediction of a target variable, given a trained predictive model. 
%
For ease of analysis and SHAP computation, we used XGBoost (eXtreme Gradient Boosting) machines \cite{chen2016xgboost} 
and trained regressors from our SC2 numeric high-level features ($529$ total) to each of the $13$ interestingness dimensions, using RMSE loss. We used the $1{,}000$ traces sampled from our RL agent, randomly selecting $80\%$ of the timesteps to get $\approx1.3\times10^6$ training instances. 
%
By looking at the mean absolute error of the models tested on the remaining $20\%$ of the data, we observed that most models achieve good prediction accuracy except for Goal Conduciveness and Incongruity,%
\footnote{Because these dimensions rely on information from multiple timesteps, a more robust model, one that makes use of past information, is likely required to provide good predictions.}
so we refrain from using these models for feature importance analysis.

\subsubsection{Global Interpretation}

\begin{figure}[!tb]
    \centering
    \includegraphics[width=\columnwidth]{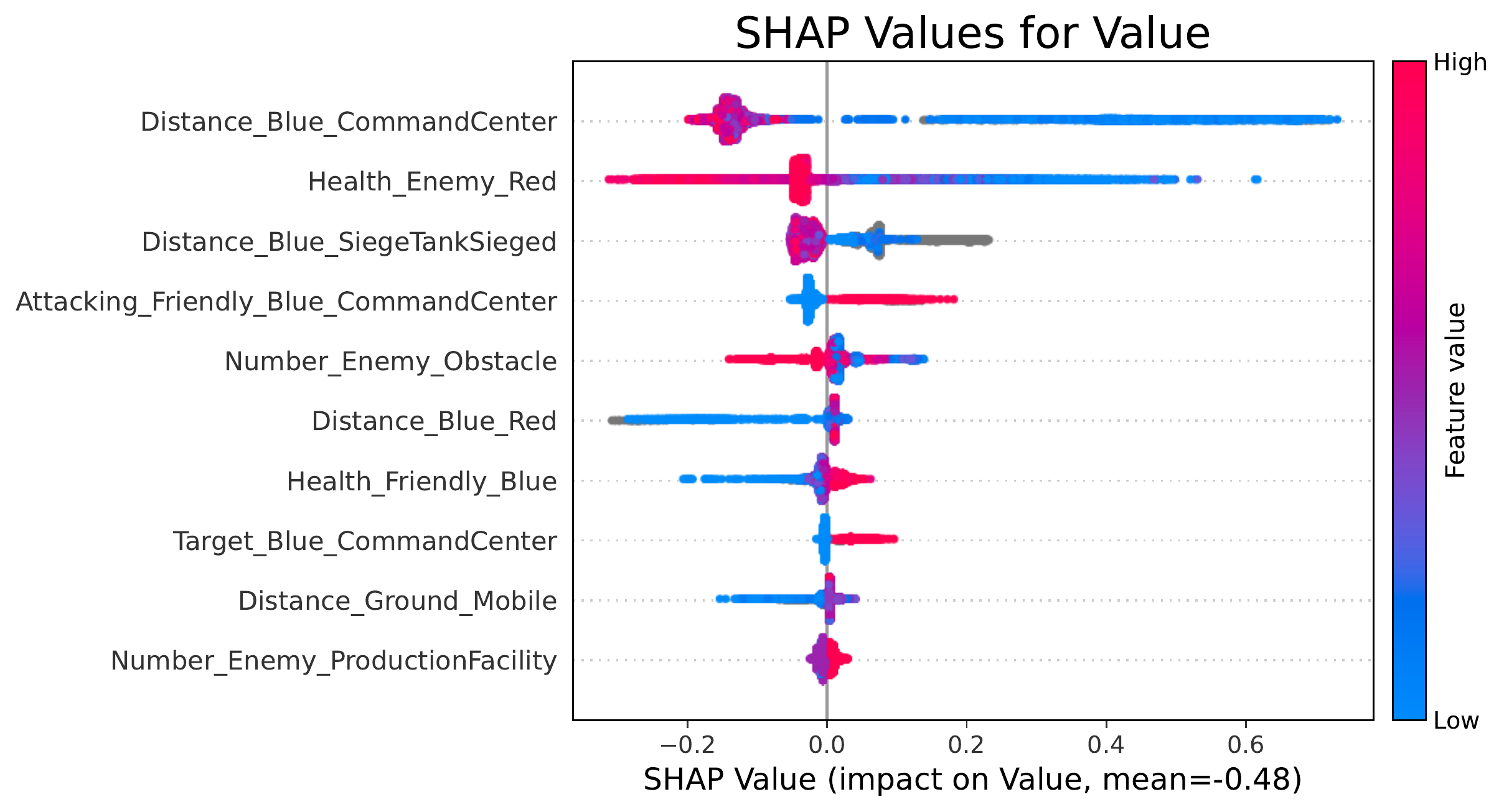}
    \caption{SHAP density plot for Value.}%
    \label{Fig:SHAPValueDensity}
\end{figure}

The goal of the global interpretation of interestingness is to understand how distinct aspects of the task influence interestingness in general. We computed the mean SHAP values %
%
%
for the test set ($\approx3\times10^5$ instances), which provides a good estimate of the features' importance for the prediction of each interestingness variable. As an example, Fig.~\ref{Fig:SHAPValueDensity} shows the SHAP density plot for the model of Value for the $10$ most predictive features (y-axis). Colored dots represent datapoints (prediction instances), stacked vertically when they have a similar SHAP value for a feature (x-axis). The color represents the corresponding feature's value, from blue (low feature values) to red (high feature values). Together, they represent how much a feature's value contributes to the prediction of Value, relative to the mean Value. As we can see, the top features show that the distance to CC (and enemy in general), the health of both forces, and whether the agent is attacking the CC, have the most impact on the predicted magnitude of Value. We also see that when the agent is attacking the CC, the closer it is, the higher the Value; and that the healthier the agent / the weaker the enemy is, the higher the Value. 

\begin{figure*}[!tb]
    \centering
    
    \begin{subfigure}[b]{0.38\textwidth}
        \includegraphics[width=\textwidth]{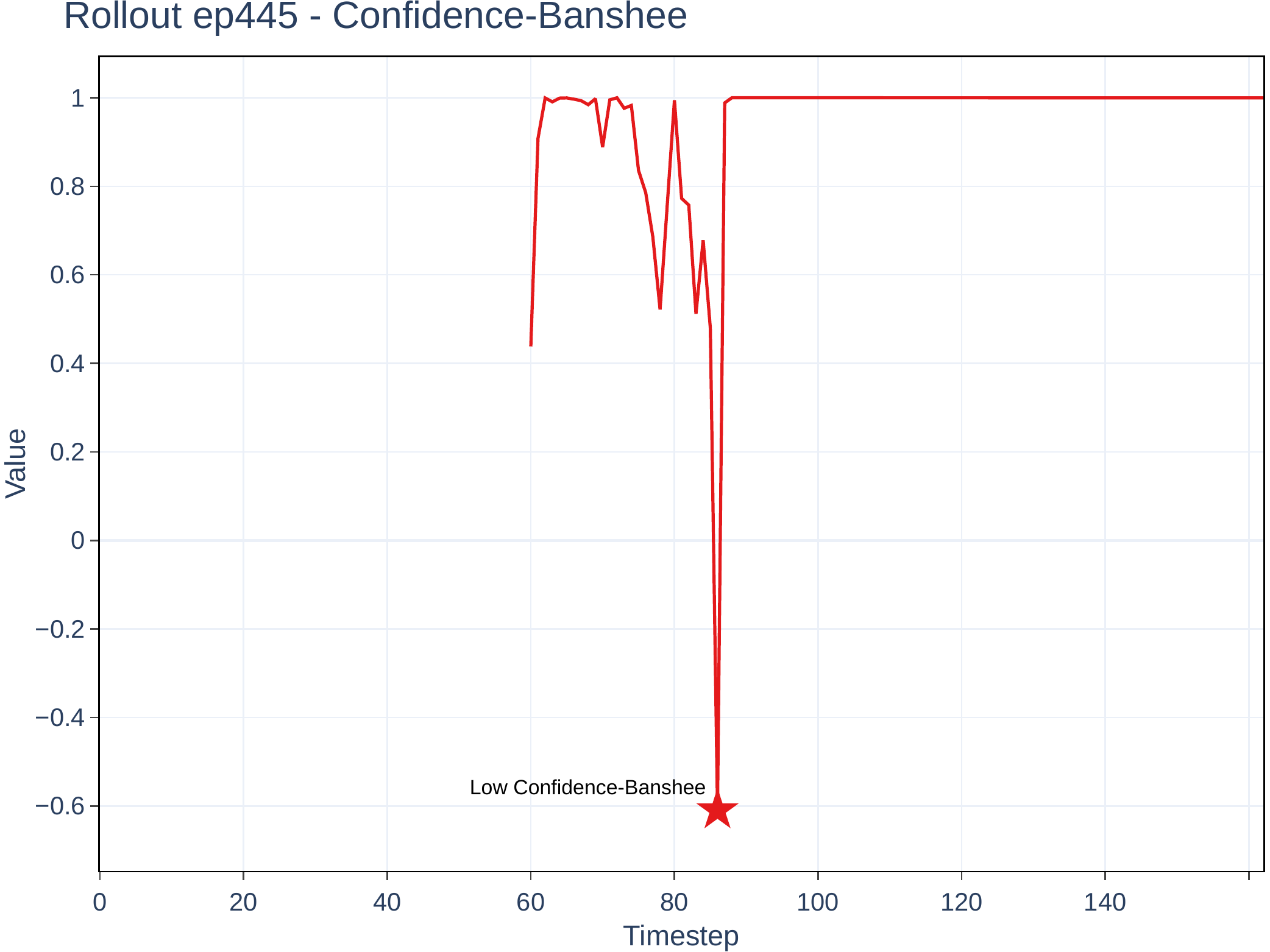}
        \caption{Confidence in using Banshees for a trace.}%
        \label{Fig:LowConfidenceBanshee}
    \end{subfigure}\hspace{10pt}%
    \begin{subfigure}[b]{0.5\textwidth}
        \includegraphics[width=\textwidth]{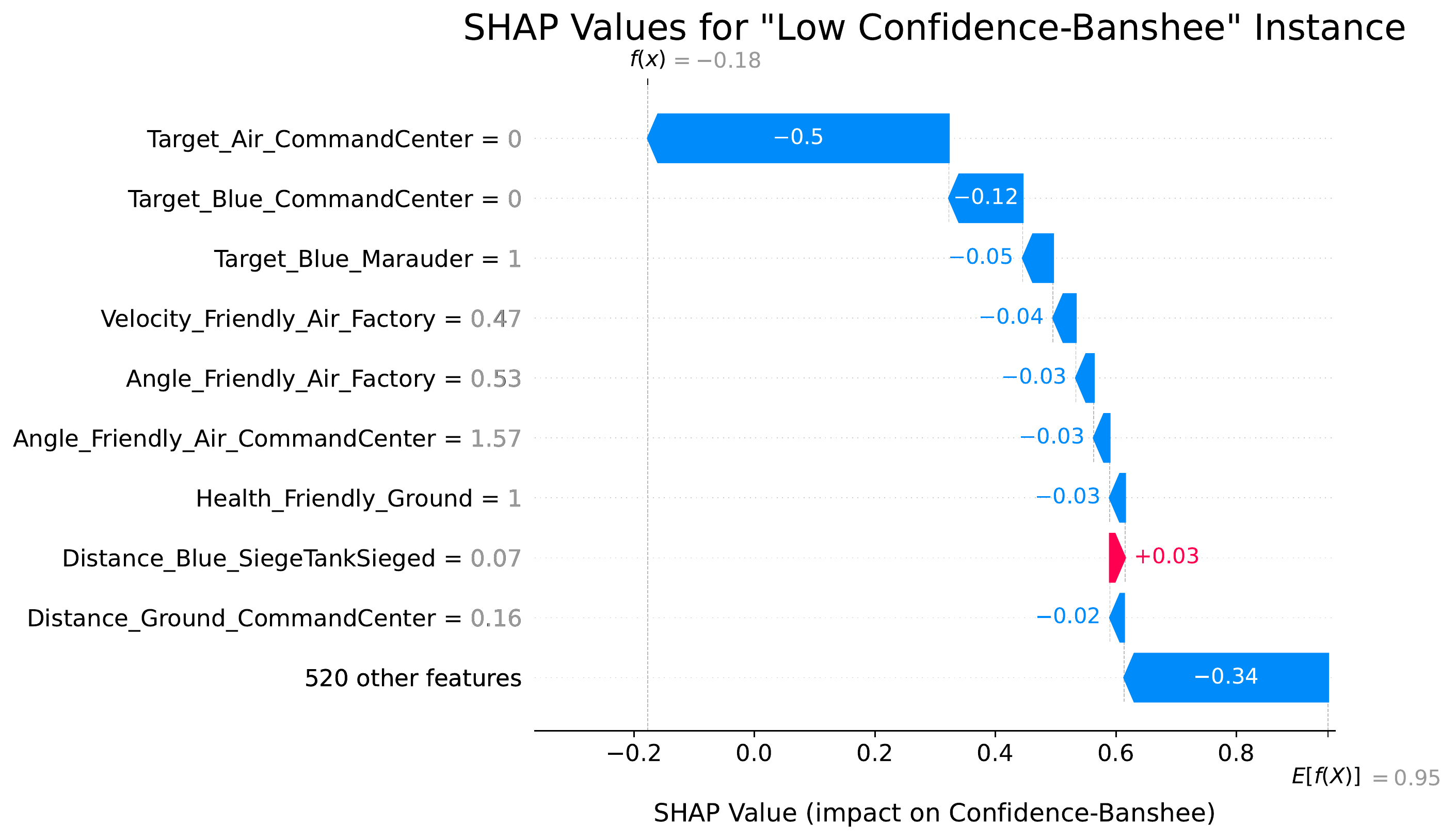}
        \caption{SHAP values for low Confidence example in (a).}%
        \label{Fig:SHAPLowConfidenceBanshee}
    \end{subfigure}
    
    
    \caption{Local interpretation example for the interestingness dimension of Confidence.}
	\label{Fig:LocalInterpretation}
\end{figure*}

\subsubsection{Local Interpretation}

Overall, global interpretation allows us to identify the task elements influencing the agent's task competency the most, indicating possible causing factors influencing the agent's capabilities and limitations. However, as discussed earlier, a good competency-aware system needs to model the situations one would expect a human operator to be curious about, ones in which a knowledge gap occurs and where further information is needed to make sense of the event \cite{hoffman2018xai}. These provide opportunities for self-explanation, requesting feedback, counterfactual reasoning, etc. 
%
As such, we use local interpretation of interestingness to understand particular ``key'' moments for the agent. 
Specifically, for each dimension, we identified the timesteps (across all the $1{,}000$ sampled traces) where the interestingness value was significantly different than the average (identified via the interquartile range method). Then, for each situation deemed ``interesting,'' we computed the SHAP values 
to identify which task elements might be responsible for the agent's high or low task competency. 

Here we illustrate the usefulness of local interpretation with one example. Fig.~\ref{Fig:LowConfidenceBanshee} plots the agent's Confidence in using Banshee for one trace, where we observe a sudden drop at around timestep $85$. From previous results, we know that the agent is usually highly confident in using Banshees. By watching the episode's replay, we see that the aerial units stopped targeting the CC and turned away from it (in the direction of the Factory) and the Blue units started targeting the Red Marauders---this information is reflected in the SHAP values for each feature, as illustrated in the waterfall plot in Fig.~\ref{Fig:SHAPLowConfidenceBanshee}, which indicates how the value of the $10$ most impactful features (y-axis) contributed to the large deviation of Confidence from the mean (x-axis). This explains why the agent suddenly had low confidence in using the Banshees: even though Banshees usually target the CC alone since they cannot be defeated, here they were ``forced'' to help the ground units. Although this particular situation may be rare, it may still lead to undesirable consequences, \eg to the loss of ground units.

By analyzing other examples of (extremely) low Confidence in using Banshees, we noted patterns in the SHAP values of those instances, as well as observed similar scenarios when watching the corresponding replays. In the future, we will automatically identify these patterns \eg by clustering the outliers based on the SHAP values for a given dimension. Additionally, end-users of our system could use the patterns to recommend alternative courses of action or remedy the agent's policy by guiding it during deployment, \eg by suggesting that the agent's ground units not engage the enemy when the agent has aerial units at its disposal. An alternative human intervention could be to retrain the agent \eg by modifying its reward function, such that when Banshees are present, ground units never engage the enemy, or, better yet, move to a safer location in the environment.

\section{Conclusions}%
\label{Sec:Conclusions}

We presented a framework for analyzing the competence of deep RL agents through the interestingness dimensions of Value, Confidence, Goal Conduciveness, Incongruity and Riskiness. Each dimension captures competence along a distinct aspect of the agent's interaction with the environment, making use of internal information by probing the models that are optimized by the algorithm during RL training. 

We conducted a computational study where we trained an RL agent in a combat task using the SC2 engine. 
Our results show that clustering traces based only on interestingness allowed the discovery of diverse behaviors, applied under distinct competency-controlling conditions. 
Furthermore, global interpretation of feature importance helped identify which (and how) task factors affect the agent's competence along each dimension. Finally, local interpretation helped determine the contribution of each task element of a particular scene---identified as having an ``abnormal'' interestingness value---which allowed gaining insights on the agent's limitations, and helps determine the measures that might need to be taken to improve the agent's performance.

Overall, our framework allows humans to have a more holistic view of an RL agent's competence, without which it is harder to understand the agent's capabilities and limitations, identify potential barriers for optimal performance, or realize what interventions need to be applied to help the agent---and its human operators and teammates---in achieving their goals. Currently, in addition to exploring ways to make better use of the interpretation mechanisms, we are developing user interfaces to facilitate the use of our tool.


\section{Acknowledgments}
This material is based upon work supported by the Defense Advanced Research Projects Agency (DARPA) under Contract No. HR001119C0112. Any opinions, findings and conclusions or recommendations expressed in this material are those of the author(s) and do not necessarily reflect the views of the DARPA.

\bibliography{22-llaama-ixdrl}

\end{document}